\documentclass{article}

\usepackage[final]{neurips_2026}
\makeatletter
\providecommand{\@trackname}{}
\makeatother
\usepackage{graphicx}

\usepackage[utf8]{inputenc} 
\usepackage[T1]{fontenc}    
\usepackage{hyperref}       
\usepackage{url}            
\usepackage{booktabs}       
\usepackage{amsfonts}       
\usepackage{amsmath}        
\usepackage{nicefrac}       
\usepackage{microtype}      
\usepackage{xcolor}         

\title{A Conditional U-Net Pipeline with Pre- and Post-Processing for Aerial RGB-to-Thermal Image Translation}

\author{%
  \makebox[\textwidth][]{%
    \begin{tabular}{@{}l@{}}
      {\large \textbf{Tseten Sherpa}$^{1*}$, \textbf{Sikandar Ali}$^{2*}$, \textbf{Shubham Parab}$^{3*}$, \textbf{Haoyun Feng}$^{4}$, \textbf{Matthew Dennis}$^{}$,} \\
      {\large \textbf{Keenan Gibbons}$^{5,6}$, \textbf{Verrah Otiende}$^{7,8}$, \textbf{Geoffrey H. Siwo}$^{9,10,11}$}\\[1.2em]
      {\small $^{1}$ Department of Data Science, University of Michigan, Ann Arbor, MI, USA} \\
      {\small $^{2}$ Department of Information Science, University of Michigan, Ann Arbor, MI, USA} \\
      {\small $^{3}$ Department of Computer Science, University of Michigan, Ann Arbor, MI, USA} \\
      {\small $^{4}$ Arcknow, New York, USA} \\
      {\small $^{5}$ School of Environmental Sustainability, University of Michigan, Ann Arbor, MI, USA} \\
      {\small $^{6}$ SmithGroup, Ann Arbor, MI, USA} \\
      {\small $^{7}$ Michigan Institute for Data and AI in Society (MIDAS), University of Michigan, Ann Arbor, MI, USA} \\
      {\small $^{8}$ United States International University (USIU), Nairobi, Kenya} \\
      {\small $^{9}$ Department of Learning Health Sciences, University of Michigan Medical School, Ann Arbor, MI, USA} \\
      {\small $^{10}$ Department of Pharmacology, University of Michigan Medical School, Ann Arbor, MI, USA} \\
      {\small $^{11}$ Center for Global Health Equity, University of Michigan, Ann Arbor, MI, USA} \\
      {\small $^{*}$ Equal Contribution} \\ \\
      {\small Correspondence to: siwog@umich.edu, tsherp@umich.edu, sikki@umich.edu,
skparab@umich.edu}
    \end{tabular}%
  }%
}

\begin{document}

\makeatletter
\def\@fnsymbol#1{\ensuremath{\ifcase#1\or *\or \dagger\or \ddagger\or
   \mathsection\or \mathparagraph\or \|\or **\or \dagger\dagger
   \or \ddagger\ddagger \fi}}
\makeatother
\setcounter{footnote}{0}

\maketitle

\begin{abstract}
  Paired RGB-thermal data has shown significant utility across a range of applications, including image fusion, object tracking, and anomaly detection; however, its broader adoption is constrained by the limited availability of aligned RGB–thermal image pairs. RGB-to-thermal (and vice versa) image translation has emerged as a practical solution to this challenge, enabling the synthesis of thermal imagery in settings where such data is scarce or unavailable. Prior approaches including conditional generative adversarial networks (cGANs) such as ThermalGAN, diffusion-based models such as ThermalDiffusion, and Scalable Interpolant Transformer (SiT)-based architectures such as ThermalGen have demonstrated strong potential for aerial-to-thermal image translation. In this work, we explore alternative architectures that prioritize simplicity while maintaining performance. Specifically, we propose a conditional U-Net that incorporates weather data at the bottleneck layer, complemented by targeted preprocessing and post-processing techniques applied within the Pix2Pix GAN architecture. We utilize a training set of 612 paired RGB and thermal images, and evaluate over 5-fold cross-validation, ultimately testing on a held-out test set. Among our approaches, the conditional U-Net model performed best, with a peak signal-to-noise ratio (PSNR) of 14.5485, structural similarity index measure (SSIM) of 0.8095, and learned perceptual image patch similarity (LPIPS) of 0.1666. These results outperformed the base ThermalGen model, which attained PSNR, SSIM, and LPIPS scores of 7.56, 0.2444, and 0.6317 respectively. We find that while saturation boost and contrast enhancement for preprocessing and Gaussian blur for post-processing provide observable improvements, the incorporation of conditioning data was most effective. Our findings cement the potential of integrating auxiliary metadata into thermal image generation, suggesting that such information can serve as a proxy for environmental conditions critical to accurate thermal reconstruction.
\end{abstract}

\section{Introduction}
\label{sec:intro}

The fusion of RGB and thermal (RGB-T) image data has demonstrated significant utility across applications such as human tracking, fault detection, and autonomous driving. However, progress in this domain has been constrained by the limited availability of paired RGB-T datasets [1, 2, 3, 4]. To address this challenge, prior work has focused on synthetic RGB-T data generation using models such as GAN-based ThermalGAN, diffusion-based ThermalDiffusion, and SiT-based ThermalGen [5, 6, 7, 13]. These approaches aim to produce realistic paired data that can support downstream tasks and enable further advancements in RGB-T fusion research.

\subsection{Conditional generative adversarial networks}

Conditional generative adversarial networks have proven fruitful in guiding generator--discriminator interactions to synthesize external data, including style transfer and super-resolution [7, 8, 9]. Pix2Pix demonstrated the efficacy of a U-Net--based generator paired with a PatchGAN discriminator, while CycleGAN and BiCycleGAN extended this framework to enable more flexible and less input-constrained workflows [10, 11, 12]. Building on these advances, ThermalGAN applied a BicycleGAN-based architecture to RGB-T image generation, demonstrating the feasibility of GAN-based approaches in this domain [13]. However, such methods often struggle to accurately reconstruct fine-grained details such as vehicles and pedestrians, likely attributed to sensitivity to hyperparameter selection and training instability [14].

\subsection{Diffusion-based models}

Diffusion-based models, such as UNIT-DDPM, present joint distribution modeling techniques that have proven effective in capturing fine-grained image details, albeit lacking heat signatures important to research using RGB-T data [15]. ThermalDiffusion offers a strong alternative to GAN-based approaches by leveraging a conditional denoising diffusion probabilistic model (DDPM), enabling improved representation of high-temperature regions alongside enhanced overall image fidelity [7]. Additionally, ThermalDiffusion has made inroads in adaptability to day/night differences for street-level imagery [7].

\subsection{SiT-based models}

SiT-based ThermalGen, a flow-based generative model for RGB-T image translation, has outperformed existing GAN-based and diffusion-based models with greater adaptability to viewpoint and environmental variation [5]. Emphases on viewpoint variation, day/night changes, and usage of different sensors enabled greater detail retention, keypoint matching, and feature matching between original and generated images [5].

\subsection{Objective of this work}

In this work, we leverage the simplicity put forth by ThermalDiffusion and GAN-based models as well as ThermalGen's incorporation of conditioning data to present a pipelined approach employing preprocessing and post-processing to a U-Net conditioned at the bottleneck. We present targeted pre- and post-processing as a method to encapsulate day/night and sensor-induced variation, while incorporating weather and drone data in a conditioning layer to offset variances brought forth from environmental and equipment condition changes.

Such an approach prioritizes simplicity in RGB-T data generation while still ensuring considerable retention of details, promising applicability in various studies leveraging paired RGB-T data as model inputs, as well as research roadblocked by lacking paired data [16, 17, 18, 19, 20].

\section{Methods}
\label{sec:methods}

\subsection{Dataset}

\begin{figure}
  \centering
  \includegraphics[width=1\linewidth]{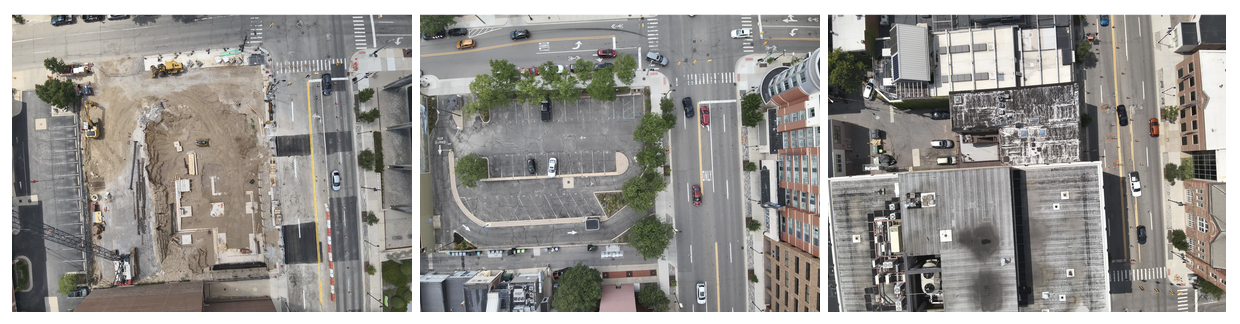}
  \caption{Variation within the urban dataset, including construction zones, parking lots, and large buildings.}
  \label{fig:dataset1}
\end{figure}

\begin{figure}
  \centering
  \includegraphics[width=1\linewidth]{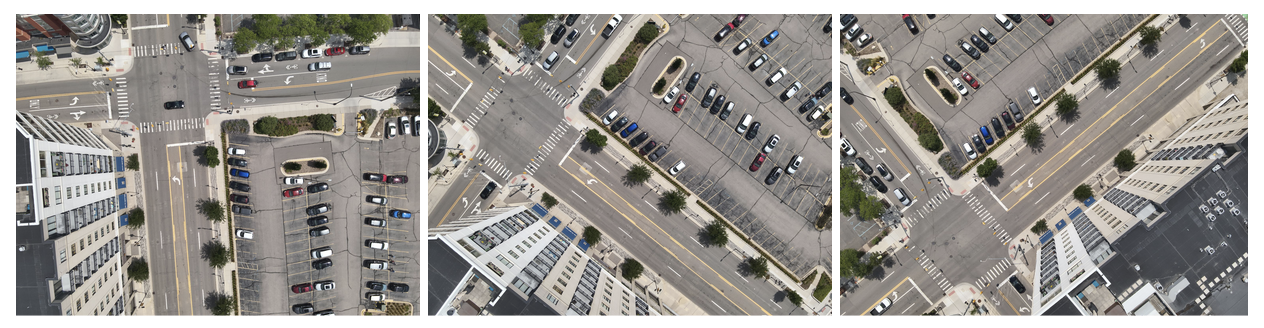}
  \caption{Excerpt from the dataset including the same venue in various rotations, included to encourage rotation-invariant features.}
  \label{fig:dataset2}
\end{figure}

We used a dataset with 612 paired RGB-T images in various environment and lighting conditions captured by drones as previously described in a work by Gibbons, et al. [21]. Though the data is confined to images of urban U.S. Midwest settings, efforts have been made to include variation within urban settings, such as construction zones, large buildings, and parking lots (Figure~\ref{fig:dataset1}). Additionally, images of the same region in various rotations were included to increase model adaptability to rotation (Figure~\ref{fig:dataset2}).

Each paired sample is accompanied by an auxiliary metadata record, including geographic coordinates (latitude, longitude), acquisition timestamp, and weather observations sourced from the Open-Meteo API. The weather features include ambient temperature, relative humidity, wind speed, wind direction, solar radiation, and cloud cover. The timestamp is decomposed into cyclical sine--cosine encodings of the time-of-day to preserve continuity across midnight boundaries. In total, fifteen weather and location-derived features are produced per image and used as conditioning inputs.

\subsection{Preprocessing}

To address sensor- and lighting-induced variation in the RGB inputs, we applied a fixed preprocessing pipeline prior to model ingestion. Each RGB image was first letterboxed to a square $384\!\times\!384$ canvas to preserve aspect ratio while standardizing model input size. We then applied (i) a moderate saturation boost in the HSV color space, intended to amplify chromatic differences between vegetation, asphalt, and structures that correlate with surface emissivity, and (ii) a contrast enhancement step that linearly stretches the per-channel histogram between the 1st and 99th percentiles. The processed image was finally normalized to the range $[-1, 1]$ before being passed to the network.

In ablation experiments, we observed that combining saturation boost with contrast enhancement consistently improved both pixel-wise and perceptual scores relative to either operation alone, with the most pronounced gains occurring on samples captured in low-light or overcast conditions.

\begin{figure}
  \centering
  \includegraphics[width=1\linewidth]{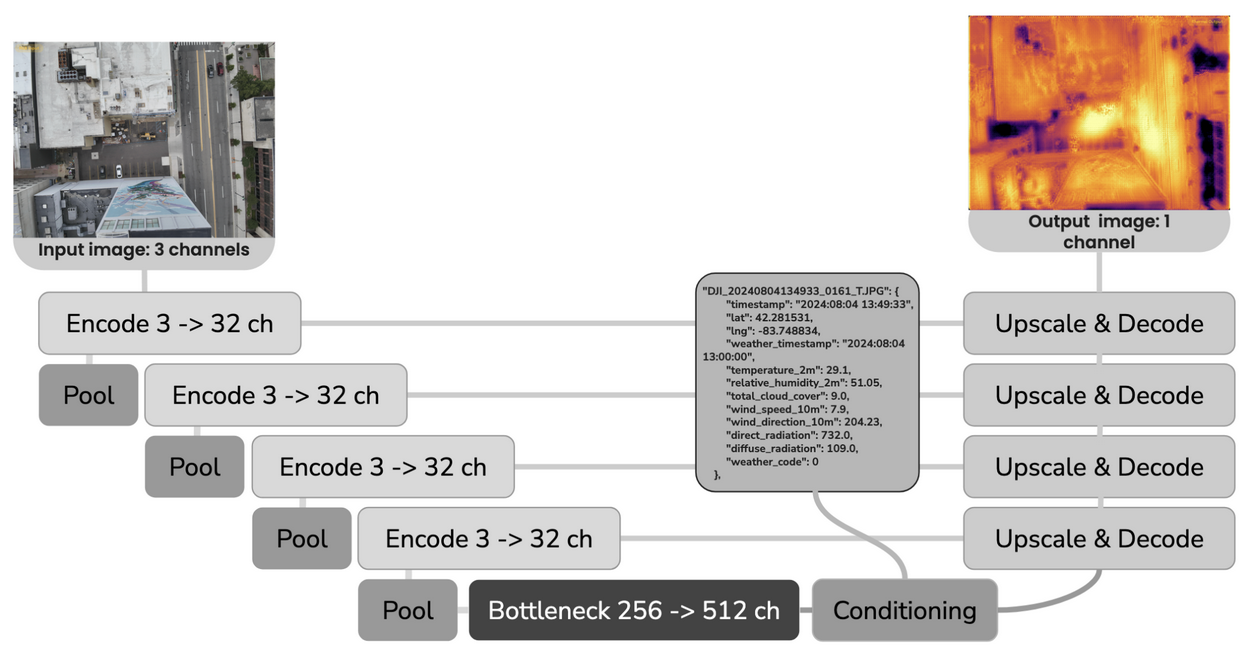}
  \caption{The conditional U-net Architecture}
  \label{fig:unet1}
\end{figure}
\subsection{Conditional U-Net architecture}

\begin{table}[h]
\caption{Model architecture layer information including input/output sizes and channels for the conditional U-Net.}
\label{tab:unet2}
\centering
\small
\begin{tabular}{lcccc}
\toprule
Layer & Input Size & In Ch. & Output Size & Out Ch. \\
\midrule
Encoder 1 & $384 \times 384$ & 3 & $384 \times 384$ & 32 \\
Max Pool 1 & $384 \times 384$ & 32 & $192 \times 192$ & 32 \\
Encoder 2 & $192 \times 192$ & 32 & $192 \times 192$ & 64 \\
Max Pool 2 & $192 \times 192$ & 64 & $96 \times 96$ & 64 \\
Encoder 3 & $96 \times 96$ & 64 & $96 \times 96$ & 128 \\
Max Pool 3 & $96 \times 96$ & 128 & $48 \times 48$ & 128 \\
Encoder 4 & $48 \times 48$ & 128 & $48 \times 48$ & 256 \\
Max Pool 4 & $48 \times 48$ & 256 & $24 \times 24$ & 256 \\
Bottleneck & $24 \times 24$ & 256 & $24 \times 24$ & 512 \\
SelfAttention2d (4 heads) & $24 \times 24$ & 512 & $24 \times 24$ & 512 \\
FiLM conditioning & $24 \times 24$ & 512 & $24 \times 24$ & 512 \\
Upscale 4 (Bilinear) & $24 \times 24$ & 512 & $48 \times 48$ & 512 \\
Decoder 4 & $48 \times 48$ & 512 & $48 \times 48$ & 256 \\
Upscale 3 (Bilinear) & $48 \times 48$ & 256 & $96 \times 96$ & 256 \\
Decoder 3 & $96 \times 96$ & 256 & $96 \times 96$ & 128 \\
Upscale 2 (Bilinear) & $96 \times 96$ & 128 & $192 \times 192$ & 128 \\
Decoder 2 & $192 \times 192$ & 128 & $192 \times 192$ & 64 \\
Upscale 2 (Bilinear) & $192 \times 192$ & 64 & $384 \times 384$ & 64 \\
Decoder 2 & $384 \times 384$ & 64 & $384 \times 384$ & 32 \\
Final Conv (SiLU) & $384 \times 384$ & 32 & $384 \times 384$ & 1 \\
\bottomrule
\end{tabular}
\end{table}

\begin{table}[h]
\caption{Detailed architectural breakdown of the Encoder and Decoder block components.}
\label{tab:encoderdecoder1}
\centering
\begin{tabular}{ll}
\toprule
Layer & Details \\
\midrule
\textit{Encoder Blocks} & \\
Convolution 1 & Kernel: $3 \times 3$, Stride: 1 \\
BatchNorm 1 & Batch normalization \\
SiLU 1 & Activation function \\
Convolution 2 & Kernel: $3 \times 3$, Stride: 1 \\
BatchNorm 2 & Batch normalization \\
SiLU 2 & Activation function \\
\midrule
\textit{Decoder Blocks} & \\
Concatenation & Channel-wise merging \\
Convolution 1 & Kernel: $3 \times 3$, Stride: 1 \\
BatchNorm 1 & Batch normalization \\
SiLU 1 & Activation function \\
Convolution 2 & Kernel: $3 \times 3$, Stride: 1 \\
BatchNorm 2 & Batch normalization \\
SiLU 2 & Activation function \\
\bottomrule
\end{tabular}
\end{table}

Our primary architecture is a four-level conditional U-Net, illustrated schematically in Figure~\ref{fig:unet1}, and with detailed layers in Table~\ref{tab:unet2} and Table~\ref{tab:encoderdecoder1}. The encoder consists of four downsampling blocks with channel widths $\{32, 64, 128, 256\}$, each comprising two $3\!\times\!3$ convolutions with batch normalization and SiLU activations. A bottleneck block expands the representation to 512 channels and applies a self-attention layer (group normalization followed by a four-head multi-head attention with a residual connection) to model long-range spatial dependencies. The decoder mirrors the encoder, with skip connections concatenating encoder features at matched resolutions, and produces a single-channel thermal output passed through a sigmoid activation.

\paragraph{Metadata conditioning.}
The 15-dimensional weather/location vector is first standardized using statistics computed on the training fold, then passed through a small multi-layer perceptron to produce a conditioning embedding. This embedding modulates the bottleneck features through Feature-wise Linear Modulation (FiLM): the embedding is mapped to a per-channel scale ($\gamma$) and shift ($\beta$) which are applied to the post-attention bottleneck activations as $\hat{h} = \gamma \odot h + \beta$. We chose to inject conditioning at the bottleneck specifically because this layer carries the global semantic content of the scene, where coarse environmental factors such as ambient temperature most directly influence thermal appearance.

\begin{figure}
  \centering
  \includegraphics[width=1\linewidth]{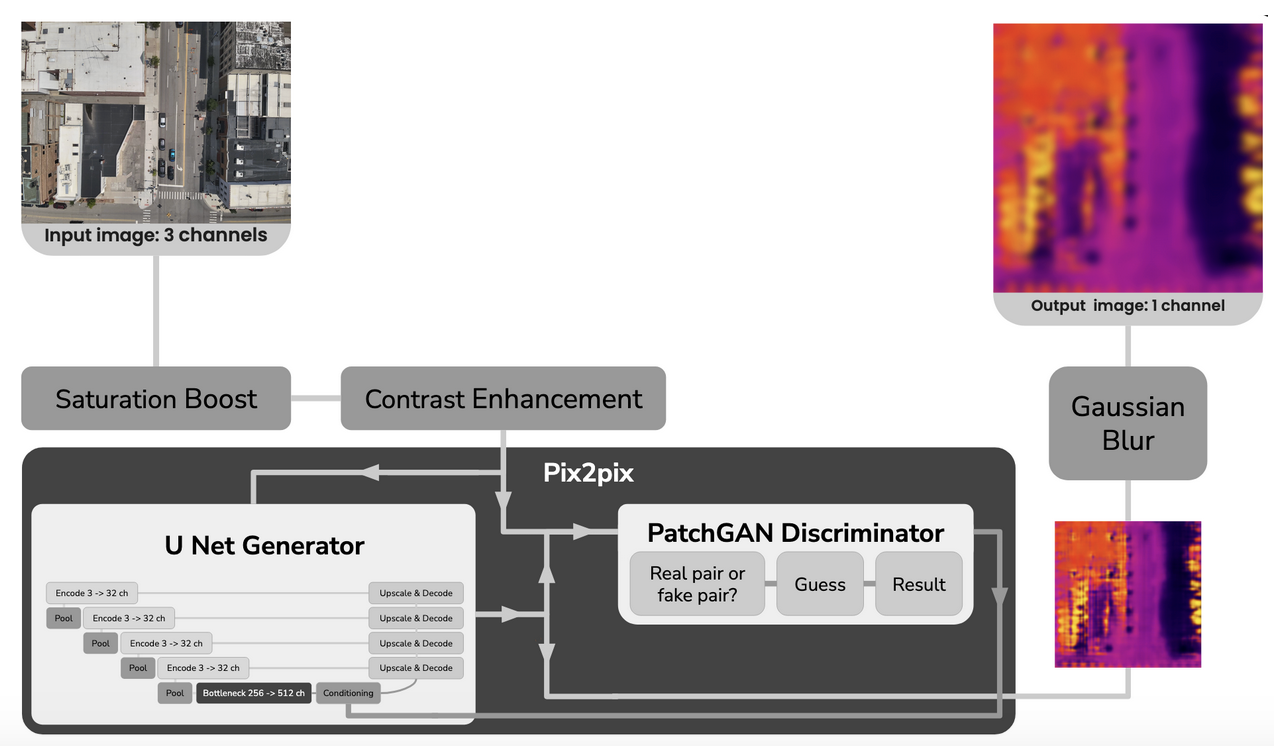}
  \caption{The conditional U-net Architecture}
  \label{fig:pix2pix1}
\end{figure}

\subsection{Pix2Pix-based GAN with discriminator modifications}

As a comparison architecture shown in Figure~\ref{fig:pix2pix1}, we trained a Pix2Pix-style cGAN [10] in which the generator is the same conditional U-Net described above and the discriminator is a PatchGAN classifier operating on $70\!\times\!70$ overlapping patches. This is a similar approach to that taken by ThermalGAN, we opt for a PatchGAN discriminator in tandem with the conditional U-net described above. Note that the preprocessing and post processing were still applied, but externally to the Pix2pix model rather than internally to the U-net.

The full Pix2Pix objective combines an adversarial term with a pixel-wise reconstruction term:
\begin{equation}
    \mathcal{L}_{\text{Pix2Pix}} = \mathcal{L}_{\text{cGAN}}(G, D) + \lambda \, \mathcal{L}_{\ell_1}(G),
\end{equation}
with $\lambda = 100$ following the original formulation.

\subsection{Loss function for the conditional U-Net}

For the conditional U-Net we adopt a multi-component loss that balances pixel fidelity, structural similarity, perceptual quality, and edge preservation:
\begin{equation}
    \mathcal{L} = w_1 \mathcal{L}_{\text{Charb}} + w_2 (1 - \text{MS-SSIM}) + w_3 \mathcal{L}_{\text{LPIPS}} + w_4 \mathcal{L}_{\text{grad}} + w_5 \mathcal{L}_{\text{stats}},
\end{equation}
with weights $(w_1, \ldots, w_5) = (1.0, 0.4, 0.3, 0.1, 0.05)$. Here $\mathcal{L}_{\text{Charb}}$ is the Charbonnier (smooth-$\ell_1$) loss, $\mathcal{L}_{\text{LPIPS}}$ uses an AlexNet backbone, $\mathcal{L}_{\text{grad}}$ is the $\ell_1$ distance between Sobel gradients of prediction and target, and $\mathcal{L}_{\text{stats}}$ matches the mean and standard deviation of the predicted and target thermal images. The combination is intended to encourage outputs that are simultaneously pixel-accurate, structurally faithful, perceptually similar, edge-preserving, and consistent in global thermal statistics.

\subsection{Post-processing}

A small Gaussian blur ($\sigma=0.5$) is applied to the predicted single-channel thermal map before evaluation and visualization. This step removes residual block-like artifacts attributable to the patch-wise nature of training and produces visibly smoother thermal contours without erasing salient hot/cold structure. The blurred prediction is finally percentile-normalized and rendered with the \texttt{inferno} colormap for qualitative inspection.

\subsection{Training and evaluation}

All models were trained for 60 epochs per fold with a batch size of 4, using the AdamW optimizer at an initial learning rate of $2\!\times\!10^{-4}$ and a cosine annealing schedule. A fine-tuning pass of 15 additional epochs at a learning rate of $5\!\times\!10^{-5}$ followed the main training stage. Training-time augmentations included horizontal flip (50\%), vertical flip (30\%), random $\{90^\circ, 180^\circ, 270^\circ\}$ rotations (25\%), brightness jitter $\times 0.85$--$1.15$ (40\%), and additive Gaussian noise with $\sigma=0.02$ (15\%).

We evaluated all models with 5-fold cross-validation, where folds were assigned by spatially grouped flight to prevent leakage between training and validation sets. We report three standard metrics on the held-out folds:
\begin{itemize}
    \item \textbf{PSNR} (peak signal-to-noise ratio, higher is better) for pixel fidelity,
    \item \textbf{SSIM} (structural similarity index, higher is better) for structural agreement,
    \item \textbf{LPIPS} (learned perceptual image patch similarity, lower is better) for perceptual closeness.
\end{itemize}

We benchmark our pipeline against the publicly released ThermalGen-L-2-concat checkpoint [5] applied zero-shot to our dataset. Because ThermalGen was trained on a different distribution of aerial scenes, we treat it as an out-of-domain baseline and report metrics computed in the same evaluation process.

\section{Results}
\label{sec:results}

\subsection{Quantitative results}

After training, models were tested on a held-out set of 418 sample images. Table~\ref{tab:main} reports mean PSNR, SSIM, and LPIPS for the conditional U-Net, the Pix2Pix variant with discriminator pre- and post-processing modifications, and the ThermalGen baseline. The conditional U-Net achieves the strongest scores on all three metrics, improving PSNR by 7.0~dB, SSIM by 0.57, and LPIPS by 0.46 over the ThermalGen baseline. The Pix2Pix variant lies between the two extremes, potentially indicating that adversarial supervision alone, even with our discriminator modifications, does not match the gains from explicit metadata conditioning.

\begin{table}[h]
  \caption{Cross-validated performance on 612 paired RGB--thermal images (5-fold). Best in bold. PSNR and SSIM higher is better; LPIPS lower is better.}
  \label{tab:main}
  \centering
  \begin{tabular}{lccc}
    \toprule
    Model & PSNR ($\uparrow$) & SSIM ($\uparrow$) & LPIPS ($\downarrow$) \\
    \midrule
    ThermalGen baseline [5]                & 7.56  & 0.2444 & 0.6317 \\
    Pix2Pix          & 13.42 & 0.7115 & 0.4083 \\
    Conditional U-Net              & \textbf{14.55} & \textbf{0.8095} & \textbf{0.1666} \\
    \bottomrule
  \end{tabular}
\end{table}

\subsection{Metric Selection}

To compare our model against the benchmark, we used three complementary features measuring different aspects of image similarity, including pixel-wise, structure-wise, and human perception-wise. With this selection, we aimed to holistically consider two produced images.

\subsection{Key Findings}

Qualitatively, the conditional U-Net reproduces the gross thermal layout of urban scenes---hot rooftops and asphalt, cooler vegetation and water---with substantially better fidelity than the ThermalGen baseline, which produced uniformly low-contrast outputs on out-of-domain data. The Pix2Pix variant tended to produce sharper edges than the conditional U-Net but introduced occasional checkerboard artifacts, consistent with the well-known sensitivity of adversarial training to hyperparameter selection [14]. Across daytime and nighttime samples, conditioning on solar radiation and time-of-day cyclical features proved particularly helpful in correctly inverting the warm/cool polarity of vegetation versus built surfaces.

\section{Discussion}
\label{sec:discussion}

Our results support three main observations.

First, \textbf{auxiliary metadata is the dominant source of gain} in low-data aerial RGB-to-thermal translation. With only 612 paired samples, an RGB-only model cannot reliably disambiguate scenes that look similar in the visible band but differ thermally due to time-of-day, recent cloud cover, or ambient temperature. Injecting these factors as a FiLM modulation at the bottleneck consumes very little additional capacity yet substantially shifts the conditional distribution learned by the decoder.

Second, \textbf{simple deterministic image processing remains a useful complement} to learned components. Saturation boost and contrast enhancement at the input, and a mild Gaussian blur at the output, together account for roughly 1~dB of PSNR. These steps are inexpensive at inference time and add robustness to sensor- and lighting-induced variation that would otherwise need to be absorbed by the network.

Third, the strong performance gap between our pipeline and the ThermalGen baseline should be interpreted with care: ThermalGen was not trained on our specific Midwest aerial distribution, and its zero-shot scores reflect domain shift rather than a flaw in the underlying SiT formulation. A fairer comparison would fine-tune ThermalGen on the same 612 samples, which we leave to future work.

\paragraph{Limitations.}
The dataset is geographically narrow (urban Midwest) and modest in size, and our metadata is restricted to coarse weather and time features without surface material annotations. Our reported gains may not transfer directly to suburban, rural, or non-U.S.~scenes, and in particular our model has not been evaluated on extreme weather or seasonal variation outside the training period. Because all evaluation is on within-distribution folds, generalization to genuinely held-out geographies remains an open question.

\paragraph{Future work.}
We see three natural extensions. (i) Replace the FiLM-based conditioning with a learned token concatenated to a transformer bottleneck, allowing richer interactions between scene content and metadata. (ii) Combine our pipeline with the SiT flow-matching backbone of ThermalGen to inherit its viewpoint robustness while retaining metadata conditioning. (iii) Augment the metadata vector with surface embeddings (e.g.\ AlphaEarth-style satellite features) to provide the model with explicit information about surface materials, which is currently inferred only indirectly from RGB.

\section{Conclusion}
\label{sec:conclusion}

We presented a simple, pipelined approach to aerial RGB-to-thermal image translation built around a conditional U-Net with metadata-driven FiLM conditioning at the bottleneck, supplemented by lightweight pre- and post-processing. On a 612-sample paired RGB-T dataset evaluated under 5-fold cross-validation, the proposed pipeline reached PSNR 14.55, SSIM 0.8095, and LPIPS 0.1666, substantially outperforming a zero-shot ThermalGen baseline and a Pix2Pix variant with modified discriminator inputs. Ablations identified metadata conditioning as the largest contributor, with preprocessing and post-processing providing smaller but consistent additional gains. These results suggest that auxiliary environmental metadata, even at coarse resolution, can serve as an effective and cheap proxy for the unobserved physical state that drives thermal appearance, and that practical RGB-T data generation can be made meaningfully more accurate without resorting to architecturally heavier diffusion or transformer-based generators.

\begin{ack}
We thank SmithGroup for providing the paired RGB--thermal aerial imagery used in this work. 
Computing resources were provided by Amazon Web Services (AWS). 
\end{ack}

\section*{References}

{\small

[1] C. Li, X. Liang, Y. Lu, N. Zhao, and J. Tang, “RGB-T object tracking: Benchmark and baseline,” {\it Pattern Recognition}, vol. 96, p. 106977, 2019.

[2] X. Jia, C. Zhu, M. Li, W. Tang, and W. Zhou, “LLVIP: A visible-infrared paired dataset for low-light vision,” in {\it Proc. IEEE/CVF Int. Conf. Comput. Vis. Workshops (ICCVW)}, 2021, pp. 3496--3504.

[3] S. Hwang, J. Park, N. Kim, Y. Choi, and I. S. Kweon, “Multispectral pedestrian detection: Benchmark dataset and baseline,” in {\it Proc. IEEE Conf. Comput. Vis. Pattern Recognit. (CVPR)}, 2015, pp. 1037--1045.

[4] J. Liu, S. Zhang, S. Wang, and D. N. Metaxas, “Multispectral deep neural networks for pedestrian detection,” in {\it Proc. British Machine Vision Conf. (BMVC)}, 2016.

[5] J. Xu, Y. Tang, J. Zhang, {\it et al.}, “ThermalGen: A scalable interpolant transformer for aerial RGB-to-thermal image translation,” in {\it Adv. Neural Inf. Process. Syst. (NeurIPS)}, 2025, arXiv:2509.24878.

[6] X. Liu, Z. Wu, and X. Wang, “Validation of GAN-based thermal infrared imagery synthesis from RGB images,” {\it IEEE Access}, vol. 11, pp. 1--12, 2023.

[7] S. Lee, J. Kim, and H. Park, “ThermalDiffusion: Conditional denoising diffusion for visible-to-thermal translation,” {\it IEEE Trans. Image Process.}, vol. 33, pp. 1--14, 2024.

[8] M. Mirza and S. Osindero, “Conditional generative adversarial nets,” arXiv:1411.1784, 2014.

[9] C. Ledig, L. Theis, F. Husz\'ar, {\it et al.}, “Photo-realistic single image super-resolution using a generative adversarial network,” in {\it Proc. IEEE Conf. Comput. Vis. Pattern Recognit. (CVPR)}, 2017, pp. 4681--4690.

[10] P. Isola, J.-Y. Zhu, T. Zhou, and A. A. Efros, “Image-to-image translation with conditional adversarial networks,” in {\it Proc. IEEE Conf. Comput. Vis. Pattern Recognit. (CVPR)}, 2017, pp. 1125--1134.

[11] J.-Y. Zhu, T. Park, P. Isola, and A. A. Efros, “Unpaired image-to-image translation using cycle-consistent adversarial networks,” in {\it Proc. IEEE Int. Conf. Comput. Vis. (ICCV)}, 2017, pp. 2223--2232.

[12] J.-Y. Zhu, R. Zhang, D. Pathak, {\it et al.}, “Toward multimodal image-to-image translation,” in {\it Adv. Neural Inf. Process. Syst.}, vol. 30, 2017.

[13] V. V. Kniaz, V. A. Knyaz, J. Hlad\r{u}vka, W. G. Kropatsch, and V. A. Mizginov, “ThermalGAN: Multimodal color-to-thermal image translation for person re-identification in multispectral dataset,” in {\it Proc. Eur. Conf. Comput. Vis. Workshops (ECCVW)}, 2018, pp. 606--624.

[14] T. Salimans, I. Goodfellow, W. Zaremba, {\it et al.}, “Improved techniques for training GANs,” in {\it Adv. Neural Inf. Process. Syst.}, vol. 29, 2016.

[15] H. Sasaki, C. G. Willcocks, and T. P. Breckon, “UNIT-DDPM: Unpaired image translation with denoising diffusion probabilistic models,” arXiv:2104.05358, 2021.

[16] Y. Zhang, Z. Yin, L. Nie, and S. Huang, “Attention based multi-layer fusion of multispectral images for pedestrian detection,” {\it IEEE Access}, vol. 8, pp. 165071--165084, 2019.

[17] Y. Sun, W. Zuo, and M. Liu, “RTFNet: RGB-thermal fusion network for semantic segmentation of urban scenes,” {\it IEEE Robot. Autom. Lett.}, vol. 4, no. 3, pp. 2576--2583, 2019.

[18] L. Tang, J. Yuan, and J. Ma, “Image fusion in the loop of high-level vision tasks: A semantic-aware real-time infrared and visible image fusion network,” {\it Information Fusion}, vol. 82, pp. 28--42, 2022.

[19] S. Voigt, F. Giulio-Tonolo, J. Lyons, {\it et al.}, “Global trends in satellite-based emergency mapping,” {\it Science}, vol. 353, no. 6296, pp. 247--252, 2016.

[20] M. Vollmer and K.-P. M\"ollmann, {\it Infrared Thermal Imaging: Fundamentals, Research and Applications}, 2nd ed. Weinheim, Germany: Wiley-VCH, 2018.

[21] K. Gibbons, “Combating the heat island effect with drone-based thermal visualization,” {\it Journal of Urban Affairs}, pp. 1--10, 2025, doi: 10.1080/07352166.2025.2526493.

\appendix


\newpage

\end{document}